\documentclass[letterpaper,twocolumn,fleqn]{article} 

\usepackage{ist}
\usepackage{bm}
\usepackage{graphicx}
\usepackage{times}
\usepackage{epsfig}
\usepackage{graphicx}
\usepackage{amsmath}
\usepackage{amssymb}
\usepackage{graphicx}
\usepackage{multirow}
\usepackage{color}
\usepackage{cite}
\usepackage{dsfont}
\usepackage[table,xcdraw]{xcolor}
\usepackage{ragged2e}
\usepackage{nccmath}
\usepackage{textcomp}
\usepackage{arydshln}
\usepackage{float}
\usepackage{hyperref}
\usepackage{tabularx}
\usepackage{gensymb}

\title{Practical cross-sensor color constancy using a dual-mapping strategy}

\author{Shuwei~Yue, Minchen~Wei*; The Hong Kong Polytechnic University, Hong Kong\newline
minchen.wei@polyu.edu.hk}

\date{} 

\begin{document} 

\maketitle

\thispagestyle{empty} 


\begin{abstract}

Deep Neural Networks (DNNs) have been widely used for illumination estimation,  which is time-consuming and requires sensor-specific data collection. Our proposed method uses a dual-mapping strategy and only requires a simple white point from a test sensor under a D65 condition. This allows us to derive a mapping matrix, enabling the reconstructions of image data and illuminants.  In the second mapping phase, we transform the re-constructed image data into sparse features, which are then optimized with a lightweight multi-layer perceptron (MLP) model using the re-constructed illuminants as ground truths. This approach effectively reduces sensor discrepancies and delivers performance on par with leading cross-sensor methods. It only requires a small amount of memory ($\sim${0.003} \textit{MB}), and takes $\sim${1} hour training on an \text{RTX3070Ti GPU}. More importantly, the method can be implemented very fast, with $\sim$0.3 \textit{ms} and $\sim$1 \textit{ms} on a \text{GPU} or \text{CPU} respectively, and is not sensitive to the input image resolution. Therefore, it offers a practical solution to the great challenges of data recollection that is faced by the industry.

\end{abstract}

\section{Introduction}

Color constancy is an ability of the human visual system that the perceived color appearance of objects remain constant under various illuminants ~\cite{gijsenij2011computational}. Digital cameras, however, do not have such an ability. Computational color constancy algorithms are developed to emulate the color constancy in the human visual system, with the key challenge being the estimation of the illuminant from a linear RAW-RGB image. In many contexts, such a challenge is similar to the real-world problem of auto white balance (AWB), which arises within the processing pipeline of digital cameras.

Traditional statistical-based methods, such as the gray-world method~\cite{buchsbaum1980spatial},  perform illuminant estimations based on individual images captured by camera sensors. They are rather simple and do not have the cross-sensor problem, but the performance is not outstanding.

Learning-based methods, such as gamut mapping~\cite{van2007edge} and color moment-based~\cite{finlayson2013corrected} methods, have also been developed for color constancy.  While they have made significant improvements compared to the statistical-based methods, recent developments in deep neural network (DNN) methods, such as~\cite{hu2017fc4}~\cite{yue2023color}~\cite{lo2021clcc}, have generally led to even better performance. These methods, however, are sensor dependent, since the relationship between the illuminants and images varies with sensors. This study focuses on DNN-based cross-sensor color constancy.

DNN-based methods, which have shown their state-of-the-art results for illuminant estimation, usually frame the problem as a regression task, learning to map input image data to illuminants as follows:
\begin{equation}
    \mathbf{L}_i={f}^{\theta}(\mathbf{Y}_i),
    \label{eq3:learning-based}
\end{equation}

\begin{figure}[!t]
  \centering
  \includegraphics[height=1.2\linewidth]{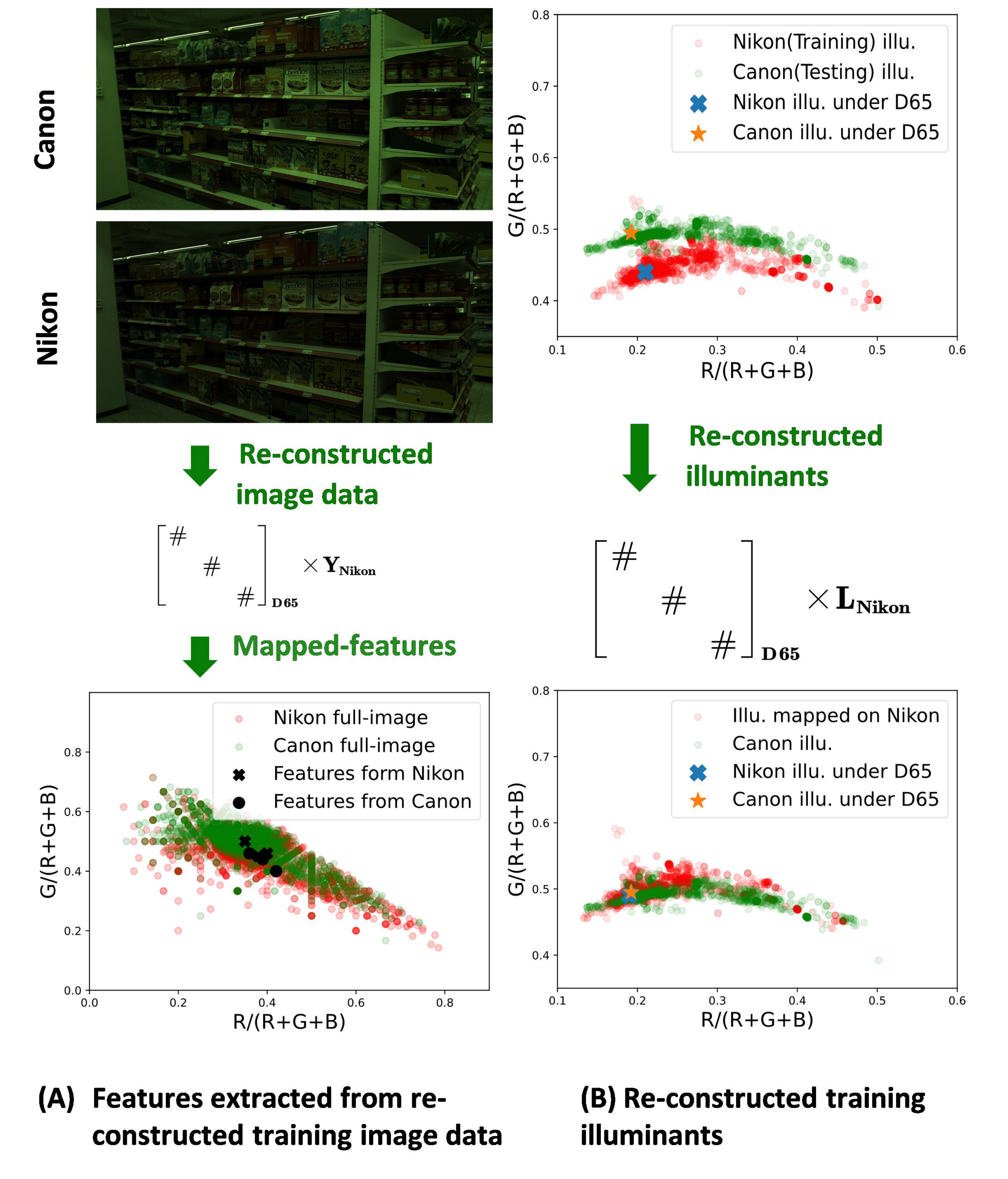}
  \caption{
Illustration of the dual-mapping strategy used in the proposed method.  (A) Illustration of the mapping of the image data captured by two camera sensors, with Nikon used as the training and Canon used as the testing, using a diagonal matrix. Such a mapping can effectively reduce the disparity of the features from the image data. (B) Illustration of the effectiveness of the mapping for illuminant distributions of the two camera sensors.}
  \label{fig:motivation}
  \vspace{-5mm}
\end{figure}

\noindent where a DNN model is trained using the linear RAW-RGB images $\mathbf{Y}_i$ and their corresponding illuminants $\mathbf{L}_i$ from a sensor-specific dataset, $i$ and $\theta$ represent the image sample index and the learning parameters, respectively.

With a great number of training datasets, DNN-based models can accurately learn the relationship between the images and ground-truth illuminants. These DNN-based models, however, need to be individually trained for each camera sensor due to the variations of the spectral sensitivity functions among different sensors and thus the variations of the image data and corresponding illuminants, as illustrated in Fig.~\ref{fig:motivation}. We denote the data of the sensor for training as a source domain ${D_s}=\left\{\mathbf{L}_{s, i}, \mathbf{Y}_{s, i}\right\}$, and the data of the test sensor as a target domain ${D_t}=\left\{\mathbf{L}_{t, i}, \mathbf{Y}_{t, i}\right\}$. Then, Eq.~\ref{eq3:learning-based} can be extended as follows:
\begin{equation}
\begin{aligned}
    &\mathbf{L}_{s, i} = {f}^{\theta_{s}}(\mathbf{Y}_{s, i})\\
    &\mathbf{L}_{t, i} = {f}^{\theta_{t}}(\mathbf{Y}_{t, i})
    \label{eq4:cross-formation}
\end{aligned}
\end{equation}
\noindent where ${f^{\theta_s}}\neq{{f^{\theta_t}}}$ due to ${D_s}\neq{{D_t}}$. Therefore, the models trained on one sensor cannot be applied directly to another sensor. Therefore, great efforts are needed to collect data, including images and illuminants (i.e., lables), for a new sensor, which becomes a great challenge for industry.
\\
\\
\noindent \textbf{Contribution}~In this paper, we present a new illumination estimation method\textemdash Dual Mapping Color Constancy (DMCC)\textemdash for cross-sensor application. 
We first calibrated a diagonal matrix $M$ using two white points captured from the training and testing sensors under D65. We separately re-constructed the image data and illuminants based on this matrix. To minimize RAW image data input variations, we map the re-constructed image data into a sparse feature space. Within this space, we observed a smaller variance between sensors compared to that of full image data, as illustrated in Fig.~\ref{fig:motivation} (A). As for the illuminant mapping between two sensors, we found that the re-constructed illuminants align well with the test sensor's illuminants, as shown in Fig.~\ref{fig:motivation} (B). This enables the generation of image data (i.e., features) and illuminant pairs that better match the test sensor's distributions, substantially reducing the need for data recollection. In summary, the performance of the DMCC method is comparable to the state-of-the-art methods, is easy to train, quick to implement, and memory-efficient, making it a practical solution to be deployed on image signal processor (ISP) chips.
\section{Prior Works}
Traditional illuminant estimation methods are beyond the scope of the discussion. Here, we only focus on DNN-based cross-sensor color constancy methods. These methods can be classified into two categories based on their practical applications: (i) model re-training-free (MRTF) methods and (ii) data re-collection-free (DRCF) methods.

\paragraph{Model Re-Training-Free (MRTF) methods}Such methods aim to develop a universal model that can be directly applied to other sensors without re-training~\cite{SIIE}~\cite{afifi2021cross}~\cite{bianco2019quasi}, or with just a little fine-tuning~\cite{xiao2020multi}~\cite{meta}~\cite{bianco2019quasi}. Such a strategy is of interest to both academia and industry since it minimizes the need for re-training and extensive data collection. The key principle behind these MRTF methods is to perform the training on diverse datasets in various domains, such RAW-RGB images captured by different sensors or even distinct color spaces~\cite{bianco2019quasi}, thus embodying multi-task learning~\cite{xiao2020multi}. This can be expressed as  $D_t\subset{D_s}$, suggesting that a well-trained model  $f^{\theta_s}$ has a great potential to have a good performance on a test set $D_t$.

Such universal models, however,  are difficult to train due to the inherent difficulty in mastering multi-domain datasets. Consequently, highly complicated DNN models are needed, which makes them difficult to deploy on ISP chips. More importantly, these models may be overfitting, if the training data are very different from the testing data, a common problem due to the wide range of camera sensors and also the differences caused by other factors (e.g., lenses). 

One of the most recent state-of-the-art methods (i.e., C5~\cite{afifi2021cross}), leverages hypernetworks~\cite{ha2016hypernetworks} and the principles of Fast Fourier Color Constancy (FFCC)~\cite{barron2017fast} to ensure reliable performance on diverse camera sensors. By incorporating hypernetworks, C5 dynamically adjusts the weightings in the model (akin to the FFCC) according to the variations of the input content, ensuring adaptability to various imaging conditions. C5's effectiveness relies on a diverse and sizable training dataset comprising labeled and unlabeled images from multiple camera sensors. Only a few images from the test camera are required for 'fine-tuning' and do not need the label information. The optimal number of images for deriving the best performance, however, varies from camera to camera. This introduces another hyperparameter, making the method more complicated and difficult for practical deployment. Moreover, complicated  data preprocessing steps, such as the log histogram operation in terms of spatial and gradient aspects, further limit its deployment.

To enlarge the training dataset size, Bianco and Cusano~\cite{bianco2019quasi} innovatively leverage sRGB images from the internet for training and directly deploy (or fine-tune) their model on the RAW-RGB testing datasets. They assume that the sRGB images that are available on the internet can generally be considered white-balanced. They then adopt a 'quasi-unsupervised' strategy to use grayscale images as input to train a DNN model to detect achromatic pixels. On one hand, such a method can relatively enlarge the size of the training dataset; on the other hand, the model can be applied to images captured by any camera. Though insightful, the heavy network and the unsatisfactory performance restrict its usage. 

Different from the previous 'learning-aware' methods, a 'color-aware' method called SIIE~\cite{SIIE} was proposed by Afifi et al. It learns an 'XYZ-like' color space in an end-to-end manner to construct the MRTF model. The assumption of the existence of an independent working space derived through a simple transformation matrix for all cameras, however, may not be valid. This can be observed from the diminished results derived based on the data from a sensor that was greatly from the training sensor. Similar to the methods discussed above, this method can also lead to overfitting.

In addition to the methods that are completely re-training-free, methods that utilize few-shot fine-tuning strategies are also available. We classify these methods into the MRTF category as well, since they also aim to create a universal model. The only difference is that minor adjustments, based on a small number of test samples, are made for a specific testing camera, which does not require very great effort for data collection. McDonagh et al.~\cite{meta} was the first to apply a meta-learning few-shot strategy (i.e., MAML~\cite{maml}) on cross-sensor color constancy problems. The method establishes initial model parameters during the meta-learning phase for optimizing the performance on unseen tasks. It makes it vital to define tasks that cover a wide range of scenarios. Specifically, the tasks are defined based on an assumption that images with a similar white point color temperature have similar dominant colors. Tuning the hyperparameters of the MAML model, however, is challenging and time-consuming due to its complexity. Inspired by this idea and the FC4~\cite{hu2017fc4} framework,  Xiao et al.~\cite{xiao2020multi} proposed a multi-task learning method (i.e., MDLCC), which includes two modules\textemdash the common feature module and the sensor-specific reweight module. Though the shared feature extractor model can effectively learn from the images captured by different camera sensors and thus increase the size of the training dataset, the method requires a high memory and becomes difficult for practical deployment. 

With the above in mind, though MRDF methods generally provide promising solutions to cross-sensor color constancy, they still have weaknesses (e.g., overfitting and complexity) for model deployment. Therefore, researchers are looking for possibilities to focus on individual testing camera sensors instead of all sensors together, and the methods are considered data recollection-free (DRCF).

\paragraph{Data Re-Collection-Free (DRCF) methods} These methods can be considered as special types of MRTF methods. Instead of aiming to train a universal model that works for all camera sensors, these methods aim to train a model for a specific camera sensor, allowing to significant reduce the workload of data re-collection. 

Such an approach directly trains a model $f^{\theta_t}$ for the test data, primarily using the source data $D_s$. An obvious drawback, in comparison to the MRTF methods, is the necessity to train a distinct model for each test sensor. Such a drawback, however, is accompanied by improvements in the model performance on the test data and also the lower likelihood of overfitting. Importantly, the DRCF methods allow a relatively lightweight model design.

Currently, there are only a few DRCF methods. One method was developed based on the Bayesian~\cite{hernandez2020multi} framework and was designed to have the ability to handle multi-task images. It uses the illuminants captured by the test camera sensors as the ground truth, trains RAW images captured by different sensors as the input data, and employs a Bayesian-based CNN framework, which leads to good performance. The necessity to collect the test illuminants, however, becomes a challenge. On one hand, these illuminants are needed for constructing the training labels. On the other hand, a comprehensive estimation of the illuminants is critical for tuning the hyperparameters of the clustering algorithms, which adds complexity to the process.  

In this article, we propose a method that only requires the white point captured by the testing camera sensor under a D65 condition, an important parameter that is always collected by camera manufacturers. Such simple data avoids the great efforts of data collection. Below, we describe the details of our proposed method and highlight the efficiency and effectiveness in addressing the challenges of the existing methods.

\section{Proposed Method}
\begin{figure*}[!ht]
\centering
\includegraphics[width=\linewidth]{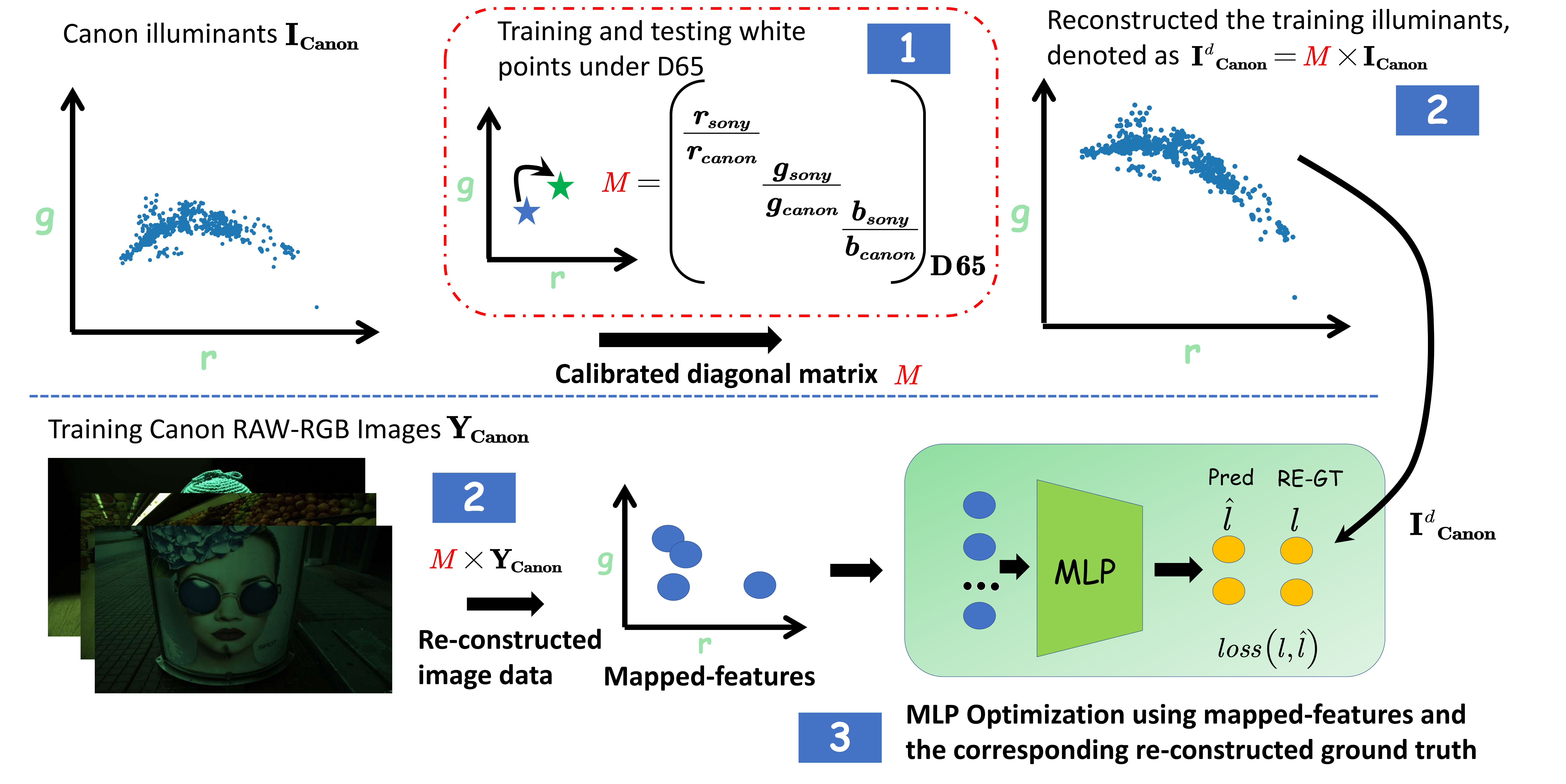}
\vspace{-2mm}
\caption{Architecture of the proposed DMCC method. It begins with the calibration step, which is used to derive the diagonal matrix $M$ using the two white points captured by the training (i.e., Canon) and testing (i.e., Sony) camera under a D65 condition. With the diagonal matrix, the training image data can be mapped using $M\times \mathbf{Y_{Canon}}$,  and the training illuminants can be mapped using $M\times \mathbf{I_{Canon}}$, labeled as $\mathbf{I^{d}_{Canon}}$. Statistical features are then extracted from the re-constructed image data. These features and the labels ${I^{d}_{Canon}}$ are used to optimize an MLP model. }
\label{fig:overview}
\vspace{-5mm}
\end{figure*}

Our proposed method (i.e., DMCC) has three steps, as illustrated in Fig.~\ref{fig:overview}. In Step 1, a diagonal matrix is derived based on the two white points, with one captured by the training and the testing camera sensors respectively, under a D65 condition, which is considered a calibration procedure. In Step 2, the diagonal matrix is used to reconstruct the image data and illuminants of the testing camera sensor. In Step 3, a multi-layer perceptron (MLP) model is trained, using the features extracted from the reconstructed image data and the reconstructed illuminants as the ground truths. Such a method can effectively reduce the differences in the data (i.e., image data and illuminants) between the training and testing camera sensors, allowing the model to be trained directly for the testing camera sensor using the data collected from the training camera sensor.

\subsection{Problem Formulation}

We propose a dual-mapping approach. It involves a calibration matrix $M$, which is derived using two white points, with one captured by the training and testing camera sensors respectively, under a D65 condition, and a feature extractor $g(\cdot)$, which is designed to align the training and testing domains. Our objective is to directly train $f^{\theta_t}$ using data pairs in the training domain, $\{\mathbf{Y}_s, \mathbf{L}_s\}$, so that data recollection is not needed for a new testing camera sensor. 

The feature extractor $g(\cdot)$ maps the reconstructed full image data $M\times\mathbf{Y}_s$ into sparse features, as illustrated in Fig.~\ref{fig:motivation} (A). It was found that the mapped features from the training and testing data are well aligned, formally:
\begin{equation}
    g(M\times\mathbf{Y}_s)\sim g(\mathbf{Y}_t).
    \label{eq: mapping-features}
\end{equation}

In addition, it was found that the distribution of the reconstructed illuminants derived using the calibration matrix and the illuminants captured by the training camera sensor is well aligned with that of the testing camera, as shown in Fig.~\ref{fig:motivation} (b). This can be expressed as:

\begin{equation} 
    M\times{\mathbf{L}_s}\sim{\mathbf{L}_t}.
    \label{eq:mapping-labels}
\end{equation}

Based on Eq.~\ref{eq: mapping-features} and Eq.~\ref{eq:mapping-labels}, it can be found that we are able to train $f^{\theta_t}$ using the pair $\{g(M\times\mathbf{Y}_s), M\times{\mathbf{L}_s}\}$, which can be symbolized as:

\begin{equation}
    \theta_t^* = \arg\min_{\theta_t} \sum_{i=1}^{n} L(M \mathbf{L}_{s,i}, f^{\theta_t}(g(M\mathbf{Y}_{s,i}))),
    \label{eq: optimiz}
\end{equation}
\noindent where $i$ is the image index, $n$ is the total number of training images, and $L(\cdot)$ is the loss function. 
It is worthwhile to point out that it is impossible to have a perfect alignment between each individual pair of the training and testing data, our proposed method is able to effectively reduce the discrepancy. Also, the efficiency of using $\{g(\mathbf{Y}), \mathbf{L}\}$ to train $f^\theta$ has been supported by our recent work~\cite{yue2023color}, in which a set of features, in terms of the chromaticities (i.e., $\left\{r, g\right\} = \left\{R, G\right\} / (R+G+B)$), is used. Specifically, the features include the maximum, mean, brightest, and darkest pixels of an image, which can be expressed as $\left\{R_{max}, G_{max}\right\}\Rightarrow{\left\{r_{max}, g_{max}\right\}}$, $\left\{R_{mean}, G_{mean}\right\}\Rightarrow{\left\{r_{mean}, g_{mean}\right\}}$, $\left\{R_b^p, G_b^p\right\} \Rightarrow \left\{r_{b}, g_{b}\right\}$($p = \text{argmax}(R_i + G_i + B_i)$), and $\left\{R_d^p, G_d^p\right\} \Rightarrow \left\{r_{d}, g_{d}\right\}$($p = \text{argmin}(R_i + G_i + B_i)$). 

In summary, the proposed DMCC method combines the feature extraction concept using $g(\cdot)$ with the reconstruction of image and illuminant data using the calibration matrix $M$, which was found effective to reduce the domain discrepancy and for dealing with the cross-sensor color constancy tasks.
\subsection{Architecture of DMCC}

The architecture of the DMCC method is improved from that of the PCC method in our recent work~\cite{yue2023color}, with modifications made to the network hyperparameters. Specifically, a grid search was conducted to determine the optimal parameters, with the number of neurons of 11 and the layer number of 5, resulting in only around 800 parameters for the network. The output of the model is the estimated illuminant chromaticities $(\hat{r}, \hat{g})$  in the 2-D chromaticity color space, with $\hat{b}$ calculated as $1-\hat{r}-\hat{g}$. Such an MLP-based network has a fast inference time, even with an unoptimized Python implementation. It only takes $\sim$0.3 ms and $\sim$1.0 ms per image on an RTX3070Ti GPU and Intel-i9 CPU, respectively. This is $\sim$25 times faster than the current fastest cross-sensor color constancy method (i.e., the C5 method). Moreover, such a fast speed is also accompanied by around $\sim$700 times fewer parameters than the C5 method. With the hardware described above, the training of the proposed DMCC model from scratch only takes less than an hour, which is considered efficient for practical deployment.

\subsection{Data Augmentation and Preprocessing}

\begin{figure}[!h]
\vspace{4mm}
\centering
\includegraphics[height=0.46\textheight]{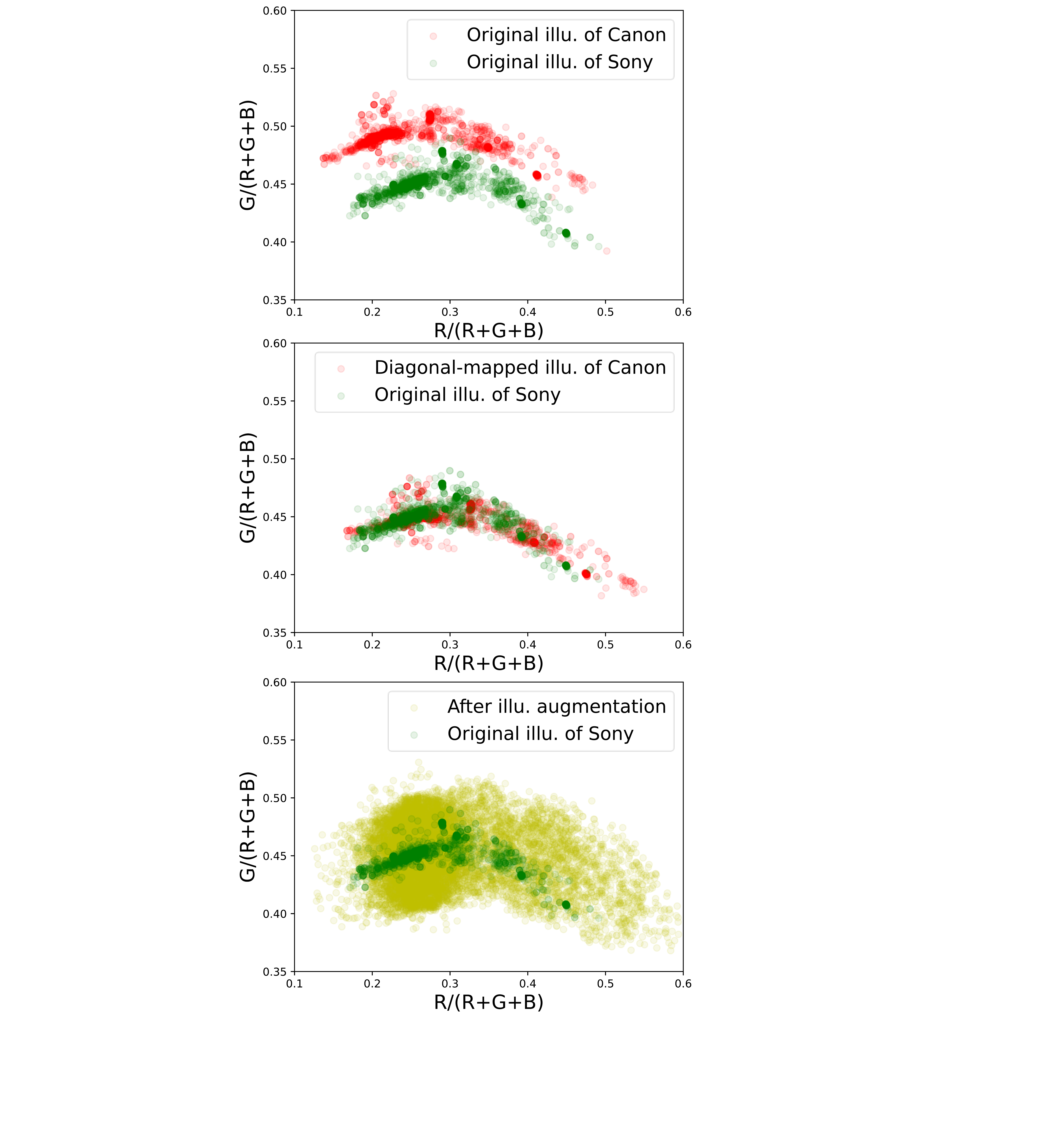}
\vspace{-14mm}
\caption{Illustration of the effectiveness of the data augmentation to cover the variations of the illuminants in the testing dataset. Top: the original distribution of the illuminants in the training and testing sets; Middle: the changes introduced by the diagonal matrix mapping; Bottom: the improved similarity between the training and testing sets after the application of data augmentation.}
\vspace{-5.5mm}
\label{fig:aug}
\end{figure}

As described above, a simple diagonal matrix is used to perform the mapping from the training to the testing sets. It is easy to understand that such a simple mapping is not able to reconstruct the testing set accurately. Thus, AWB-Aug~\cite{yue2023color} is employed to perform the data augmentation, which involves an illuminant enhancement strategy. Specifically, uniform sampling is performed around the illuminant in the chromaticity space, with the illuminant positioned at the center of the circle. The radius of the circle, a hyperparameter, is set to 0.05, which was found to produce stable results, as shown in Fig.~\ref{fig:aug}.

In the experiment, linear RAW RGB images, with the calibration labels and black level subtracted, were used. Also, over-saturated and darkest pixels, as described in~\cite{yue2023color}, were clipped. Moreover, since the model is based on sparse features and is resolution-independent, the images were resized to $64\times64\times3$ and normalized for fast processing, 

\subsection{Implementation Details}
The proposed DMCC method adopts the traditional angular error between the estimated illuminant $\hat{\pmb{\ell}}$ and the ground truth illuminant ${\pmb{\ell}}$ with a regularization as the loss function:
\begin{equation}
    \pmb{L}(\theta)=cos^{-1}\left(\frac{\pmb{\ell}\odot{\hat{\pmb{\ell}}}}{\left\|\pmb{\ell}\right\|\times{\left\|\hat{\pmb{\ell}}\right\|}}\right) + \lambda ||\theta||_1,
\end{equation}
where $\odot$ represents the inner products and $cos^{-1}(\cdot)$ is the inverse of a cosine function. L1 regularization is employed to adjust the training parameters $\theta$ to avoid overfitting, and $\lambda$ is the regularization weighting factor of $10^{-5}$.

The DMCC framework, constructed with PyTorch and integrated with CUDA support, uses the Adam optimizer~\cite{kingma2014adam} for training, in conjunction with He initialization~\cite{He2015Delving}. We utilize a batch size of 32 over $10,000$ epochs with a learning rate of $7\times10^{-3}$. In addition, we apply a cosine annealing strategy~\cite{sgd} to adjust the learning rate and employ an early stopping strategy to save the best-performing model throughout the training process.

\begin{table*}[!ht]
\centering
\caption{Table 1. Summary of the performance of various methods, in terms of angular errors, on the INTEL-TAU datasets, together with the processing time and parameter size. The results of the Gray-World, White Patch, Shades-of-Gray, and Cheng-PCA were extracted from ~\cite{intel-tau}, and those of the Quasi-Unsupervised, SIIE, FFCC, C5, and MDLCC were extracted from~\cite{afifi2021cross} and~\cite{xiao2020multi}.} 
\label{TAU-results}
\vspace{5mm}
\scalebox{1.0}{
\begin{tabular}{l|ccccc:cc}
\textbf{\begin{tabular}[c]{@{}l@{}}\textbf{INTEL-TAU Dataset}\\\textbf{Method} \end{tabular}} & \textbf{\begin{tabular}[c]{@{}c@{}}Best \\ 25\%\end{tabular}} & \textbf{Mean} & \textbf{Med.} & \textbf{Tri.} & \textbf{\begin{tabular}[c]{@{}c@{}}Worst \\ 25\%\end{tabular}} & \textbf{Time(ms)} & \textbf{Size(MB)}\\
\hline
Gray-world \cite{buchsbaum1980spatial}& 0.9 & 4.7 & 3.7 &  4.0 & 10.0 &- &-\\
White-Patch \cite{white-patch}& 1.1 & 7.0 & 5.4 & 6.2 & 14.6 & - & -\\
Shades-of-Gray~\cite{finlayson2004shades}& 0.7 & 4.0 & 2.9 & 3.2 & 9.0 &- &- \\
Cheng-PCA~\cite{cheng2014illuminant} & 0.7 & 4.6 & 3.4 & 3.7 & 10.3 &- &- \\

Quasi-Unsupervised CC~\cite{bianco2019quasi} & 0.7 & 3.7 & 2.7 & 2.9 & 8.6 & 90 & 622\\
SIIE~\cite{SIIE} & 0.7 & 3.4 & 2.4 & 2.6 & 7.8 &35 & 10.3 \\
FFCC~\cite{barron2017fast} & 0.7 & 3.4 & 2.4 & 2.6 &8.0 & 23 & 0.22 \\ 
MDLCC~\cite{xiao2020multi} &- &- &- &- & - &25 &6\\
C5(m=7)~\cite{afifi2021cross}& 0.5 & 2.6 & 1.7 & - & 6.2 & 7 & 2.09\\
C5(m=1)~\cite{afifi2021cross} & 0.7 & 3.0 & 2.2 & - &6.7  & 7 & 2.09 \\
\hdashline
DMCC(Ours) & 0.7 & 3.0 & 2.3 & 2.2 & 6.8 &0.3  & 0.003\\
\end{tabular}
}
\vspace{5mm}
\end{table*}


\begin{figure*}[!ht]
\vspace{-2mm}
\centering
\includegraphics[width=\linewidth]{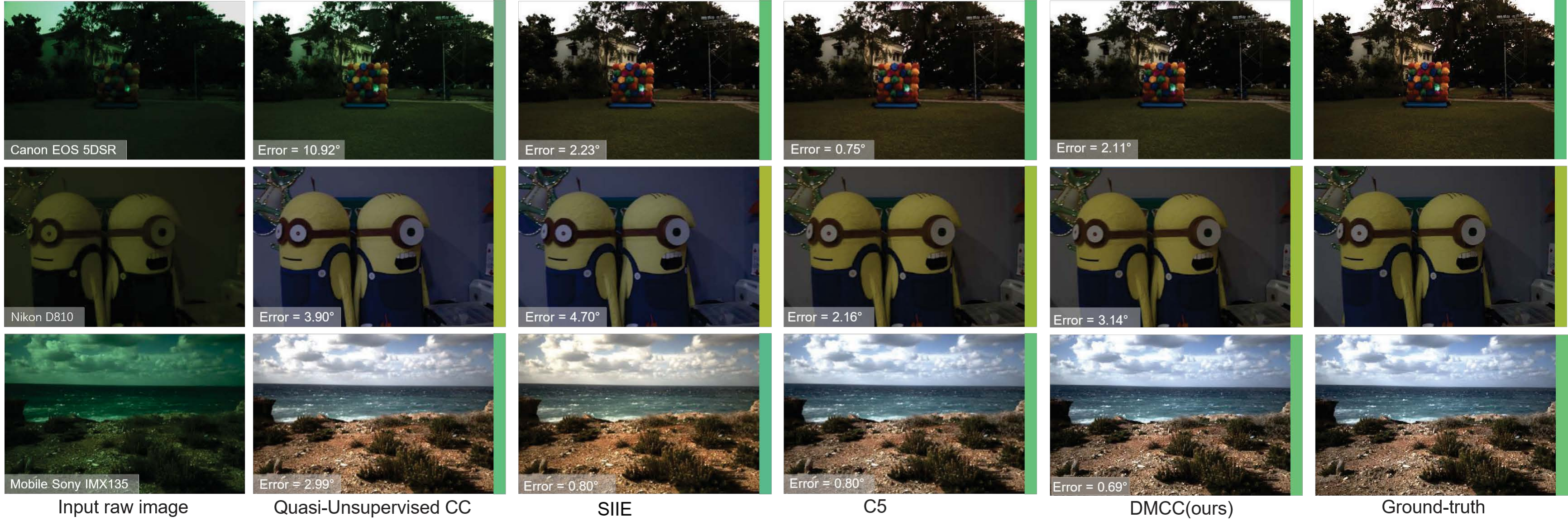}
\caption{Examples of the images processed using the proposed DMCC method, and other methods extracted from ~\cite{afifi2021cross}.}

\label{fig:results-qualitative}      
\vspace{-3mm}
\end{figure*}

\section{Experimental Results}

The proposed DMCC method was validated on the INTEL-TAU~\cite{intel-tau} dataset, which includes 7,022 images captured by three different cameras (i.e., Canon 5DSR, Nikon D810, and Mobile Sony IMX135). We followed the cross-sensor training and testing strategies, aligning with the INTEL-TAU strategy for a fair comparison. Five metrics, such as the mean, median (Med.), trimean (Tri.), the mean of the smallest 25\% (Best~25\%), and the mean of the largest 25\% (Worst~25\%) of the angular errors between the estimated and the ground-truth illuminants, were used to show the performance.

The average results from the three experiments are shown in Table~\ref{TAU-results}. It can be observed that the DMCC method has much better performance than the statistical-based methods, and also has comparable performance to the C5 method (m=1 or 5, where m is the number of image samples utilized from the test camera sensor). Fig.~\ref{fig:results-qualitative} shows the images from the various methods, which directly shows the performance of the DMCC method.

\section{Discussion}
In addition to using a diagonal matrix, a full matrix was also used to see whether it can lead to a better performance. It was found that the diagnoal matrix derived at 6500 K had a better performance, reducing the mean of the angular error by around 1\degree. This was likely due to the prevalence of daylight conditions in most scenes. Furthermore, the bad performance of the full matrix was likely due to the linear transformation errors across a wide range of CCT levels.

\section{Conclusion}
We propose a method (i.e., DMCC) using a dual-mapping strategy for the problem of cross-sensor illuminant estimation. The method performs the training on the testing camera sensor, which differs from the conventional methods that heavily rely on extensive data collection and complicated modeling. Specifically, the first mapping employs a diagonal matrix, which is derived from white points captured by the training and testing camera sensors under a D65 condition, to reconstruct the image data and illuminants. Then, the second mapping transforms the reconstructed image data into sparse features. These features, along with the reconstructed illuminants serving as the ground truths, are used to optimize a lightweight MLP model. The proposed method results in a good performance, which is comparable to the state-of-the-art solutions. More importantly, it is compact with only $\sim$0.003 MB parameters, requiring just 1/700 of the memory size of its advanced counterparts. It also achieves a rapid inference time of $\sim$0.3 ms on a GPU, about $\sim$25 times faster. In summary, the method provides a practical and efficient solution to AWB for practical deployment.





\end{document}